\documentclass[letterpaper, 10 pt, conference]{ieeeconf}  

\IEEEoverridecommandlockouts                              

\usepackage{graphics} 
\usepackage{epsfig} 
\usepackage{mathptmx} 
\usepackage{times} 
\usepackage{amsmath} 
\usepackage{amssymb}  

\usepackage{float}
\usepackage{multirow}
\usepackage{array}

\usepackage{graphicx}
\usepackage{booktabs}
\usepackage{subcaption}

\usepackage{setspace}
\usepackage[style=ieee, maxcitenames=2, mincitenames=1,natbib=true,dashed=false]{biblatex}
\title{\LARGE \bf
Pedestrian Crossing Action Recognition and Trajectory Prediction \\with 3D Human Keypoints
}

\author{Jiachen Li$^{1*}$, Xinwei Shi$^{2\dagger}$, Feiyu Chen$^{2\dagger}$, Jonathan Stroud$^{2\dagger}$, Zhishuai Zhang$^2$, Tian Lan$^2$, Junhua Mao$^2$,\\ Jeonhyung Kang$^2$, Khaled S. Refaat$^2$, Weilong Yang$^2$, Eugene Ie$^2$ and Congcong Li$^2$
\thanks{* Work done during Jiachen's internship at Waymo}
\thanks{$^\dagger$ Equal contributions}
\thanks{$^1$ Stanford University, {\tt\footnotesize jiachen\_li@stanford.edu}}
\thanks{$^2$ Waymo, {\tt\footnotesize \{xinweis, junhuamao\}@waymo.com}}
}

\begin{document}

\maketitle
\thispagestyle{empty}
\pagestyle{empty}

\begin{abstract}
Accurate understanding and prediction of human behaviors are critical prerequisites for autonomous vehicles, especially in highly dynamic and interactive scenarios such as intersections in dense urban areas.
In this work, we aim at identifying crossing pedestrians and predicting their future trajectories.
To achieve these goals, we not only need the context information of road geometry and other traffic participants but also need fine-grained information of the human pose, motion and activity, which can be inferred from human keypoints. 
In this paper, we propose a novel multi-task learning framework for pedestrian crossing action recognition and trajectory prediction, which utilizes 3D human keypoints extracted from raw sensor data to capture rich information on human pose and activity.
Moreover, we propose to apply two auxiliary tasks and contrastive learning to enable auxiliary supervisions to improve the learned keypoints representation, which further enhances the performance of major tasks.
We validate our approach on a large-scale in-house dataset, as well as a public benchmark dataset, and show that our approach achieves state-of-the-art performance on a wide range of evaluation metrics. The effectiveness of each model component is validated in a detailed ablation study.

\end{abstract}

\section{INTRODUCTION}
Accurately understanding the behaviors of surrounding traffic participants and predicting their future motions play a significant role in autonomous driving to ensure safe and efficient interactions between the self-driving vehicle and surrounding agents in the scene.
Among various types of traffic participants, pedestrians are often exposed to and involved in traffic accidents and fatalities, especially when they are crossing the street in dense urban areas \cite{rasouli2019autonomous}.
Therefore, in order to achieve safe navigation and deployment of autonomous vehicles in populated and highly interactive environments, it is essential to capture informative cues about the actions, intentions, and motions of pedestrians. 

\begin{figure}[!tbp]
	\centering
	\includegraphics[width=\columnwidth]{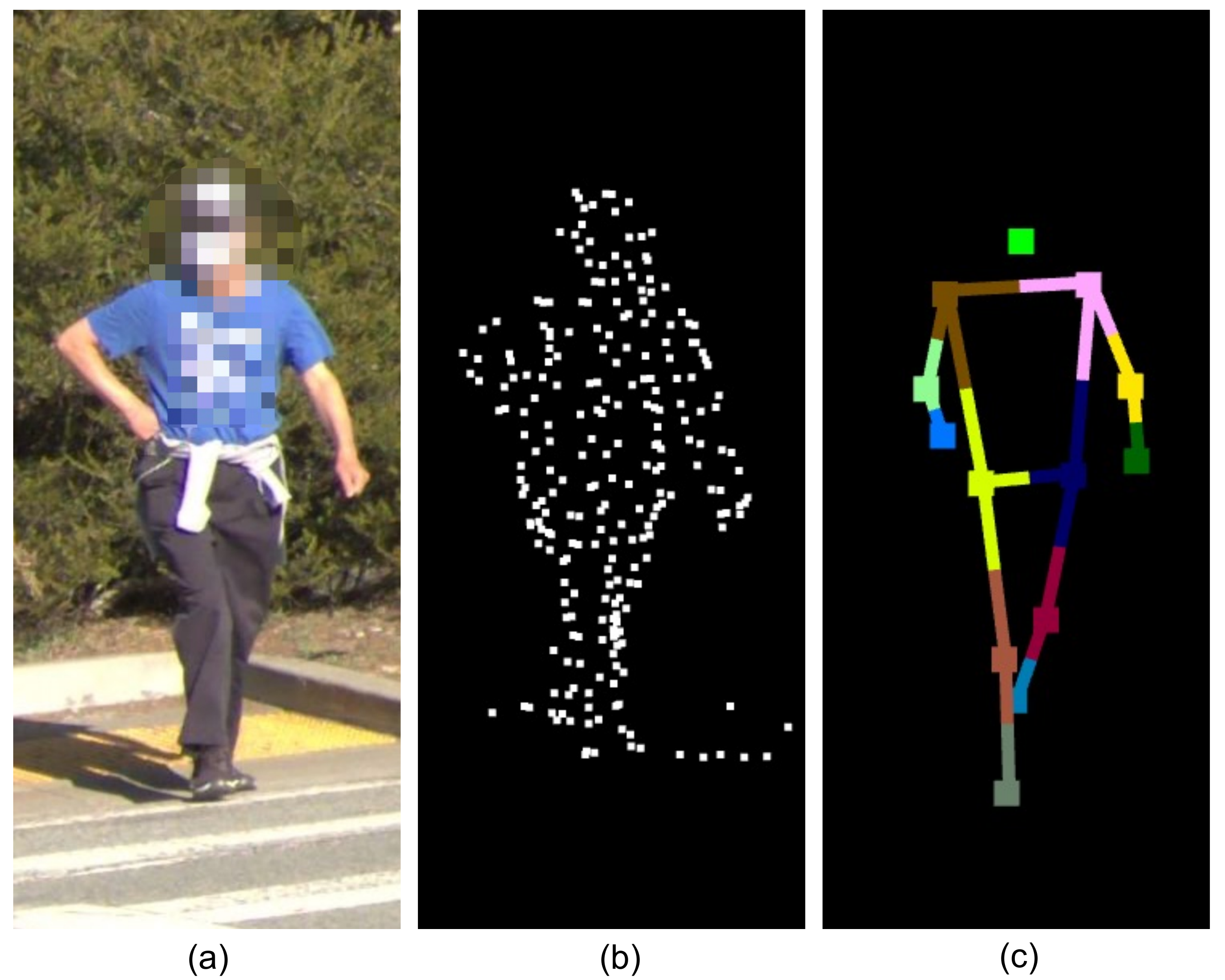}
	\caption{A comparison between different forms of appearance information: (a) 2D image patch; (b) 3D point cloud; (c) 3D human keypoints inferred from (b).} 
	\vspace{-0.5cm}
	\label{fig:teaser}
\end{figure}

Specifically, we aim to recognize whether pedestrians are actively crossing the street (crossing action recognition), and predict their future trajectories (trajectory prediction).
The former is typically formulated as a binary classification problem \cite{liu2020spatiotemporal,yao2021coupling,yau2021graph}, and heavily depends on the appearance features that provide rich information about the pedestrians' poses and activities. An illustrative comparison between typical forms of appearance information is provided in Fig. \ref{fig:teaser}.
The latter aims to predict plausible future trajectory hypotheses which may be diverse with different target locations \cite{zhao2020tnt,mangalam2020not}.
It heavily depends on the history trajectories due to the continuity of human motions.
Both tasks also require knowledge about roadgraph for a better understanding of the environmental context.
In previous work, these two tasks have mostly been studied separately. However, since both tasks require extracting features from some common inputs, they can potentially enhance each other by being integrated into a unified framework.

Existing methods of crossing action recognition focus on extracting the appearance features of pedestrians either based on 2D images or video frames. Some of them further infer human keypoints in the image plane to provide a more compact representation.
However, the 2D human keypoints that are used as the input of keypoints encoder vary as the view angle changes.
Most existing trajectory prediction approaches only utilize the low-dimensional state information (e.g., position, velocity) as history observations without knowledge about the appearance of target pedestrians \cite{zhao2020tnt,li2021rain,mohamed2020social,zhou2022grouptron,salzmann2020trajectron++,cao2021spectral}. They lack the appearance information that can provide useful cues for motion prediction, especially when the pedestrians are performing special activities (e.g., bending down, waving hands) or interacting with objects (e.g., pushing a cart, or riding a scooter).
Some recent methods take advantage of raw sensor data (e.g., point cloud) directly \cite{laddha2021mvfusenet,djuric2020multixnet}, but it may include redundant or sparse information.
3D human keypoints can provide a compact and clean representation for extracting spatio-temporal features of pose and activity.

To the best of our knowledge, we are the first to investigate the utilization of 3D human key points to leverage pedestrian appearance information for crossing action recognition and trajectory prediction in a unified multi-task learning framework.
In contrast to the 2D information, 3D human keypoints contain the appearance information that can be transformed to be invariant of view angle, and in the same coordinate system for trajectory prediction. 
Therefore, we adopt an effective keypoints encoder and design three auxiliary supervisions inspired by self-supervised learning to encourage better keypoints representation learning. 
More specifically, we generalize the pretext tasks in the context of self-supervised learning that are designed to learn spatial patterns of images to capture temporal patterns of the keypoints sequences \cite{jing2020self,liu2021self,noroozi2016unsupervised,misra2020self,carlucci2019domain,jaiswal2021survey,chen2020simple,khosla2020supervised}.

The contributions of this paper are as follows. We propose a novel multi-task learning framework for pedestrian crossing action recognition and trajectory prediction, which leverages 3D human keypoints to capture human poses and activities effectively. To our knowledge, we are the first to demonstrate the effectiveness of utilizing 3D human keypoints for these tasks by capturing more fine-grained and complete features about pose and activity. We also propose to employ two auxiliary tasks and contrastive learning to enable additional supervisions for keypoints representation learning, which also enhance the performance of primary tasks. We validate the proposed method on a large-scale in-house dataset and a public benchmark dataset, which achieves state-of-the-art performance in terms of a wide range of evaluation metrics. 

\section{Related Work}

\textbf{Crossing Action Recognition}:
Pedestrian crossing action recognition is widely studied since it is crucial for safe interaction with pedestrians and efficient decision making for autonomous vehicles, which is typically formulated as a binary classification problem (i.e., crossing or non-crossing) \cite{liu2020spatiotemporal,girase2021loki,yao2021coupling,singh2021multi,yau2021graph,rasouli2020pedestrian}.
Many existing public datasets (e.g., PIE \cite{rasouli2019pie} and JAAD \cite{kotseruba2016joint}) assume that only 2D camera information is available, so existing approaches can only leverage the frontal-view images or videos without 3D information of the pedestrians. 
Thus, the appearance features can only be extracted visually from image patches inside the pedestrian bounding box.
Some approaches further use pose estimation models \cite{cao2019openpose} to obtain 2D human keypoints from images to provide richer cues to improve performance \cite{yang2021predicting}.
However, some keypoints may be missing due to overlap or occlusion in the 2D plane.
Since many autonomous vehicles are equipped with sensors (e.g., LiDAR) that capture 3D information, we propose to leverage 3D human keypoints to capture more fine-grained features about pose and activity.

\textbf{Trajectory Prediction}:
Many research efforts have been devoted to trajectory prediction for pedestrians and vehicles in highly dynamic and interactive scenarios. Many existing methods utilize the high-level history state information (e.g., position, velocity) and the context information (e.g., roadgraph/map, context agent trajectory) to forecast future state sequences \cite{li2020evolvegraph,gao2020vectornet,choi2021shared,ma2021continual,zhao2020tnt,alahi2016social,li2021rain,mohamed2020social,chai2019multipath,li2019conditional,salzmann2020trajectron++,sun2022interaction}.
There are two widely used ways to represent the roadgraph information: (a) rasterized top-down view images \cite{chai2019multipath,li2020evolvegraph,toyungyernsub2022dynamics}; and (b) roadgraph vectors \cite{gao2020vectornet,zhao2020tnt}.
In order to model the interactions between entities, different feature aggregation techniques are employed such as social pooling \cite{alahi2016social}, attention mechanisms \cite{li2021rain}, and message passing across graphs \cite{mohamed2020social}.
However, these models cannot capture any appearance information of traffic participants due to the lack of sensor data (e.g., point cloud), which may lose fine-grained cues that are important for predicting their future motions.
Recent works validated the advantages of using raw sensor data \cite{laddha2021mvfusenet,djuric2020multixnet}.
In contrast, we investigate the effectiveness of additionally leveraging the appearance information in the form of 3D human keypoints inferred from laser points in trajectory prediction.

\textbf{Human Keypoints Encoding}:
Extracting informative features from human keypoints is crucial for human action recognition and motion prediction \cite{ma2022multi,wei2020his,li2020dynamic,walker2017pose,guo2019human,chiu2019action,li2018convolutional,martinez2017human,gopalakrishnan2019neural}.
Deep convolutional networks are applied to human keypoints detected in 2D images or videos to extract visual features \cite{li2018convolutional,walker2017pose}.
For human keypoints in 3D space, recurrent neural networks are usually employed to capture temporal dynamics \cite{gopalakrishnan2019neural,martinez2017human,guo2019human}. However, it is hard for the recurrent units to capture the spatial relations between the joints in the human skeleton.
As the development of graph representation learning, graph neural networks are adopted to extract spatio-temporal feature representations from a spatio-temporal graph constructed by the human keypoints sequence, where the nodes represent human joints and edges represent the bones between joints \cite{li2020dynamic,yan2018spatial}.
In this work, we employ the Spatio-Temporal Graph Convolutional Network proposed in \cite{yan2018spatial} as the backbone of our keypoints encoder.

\begin{figure*}[!tbp]
	\centering
	\includegraphics[width=0.9\textwidth]{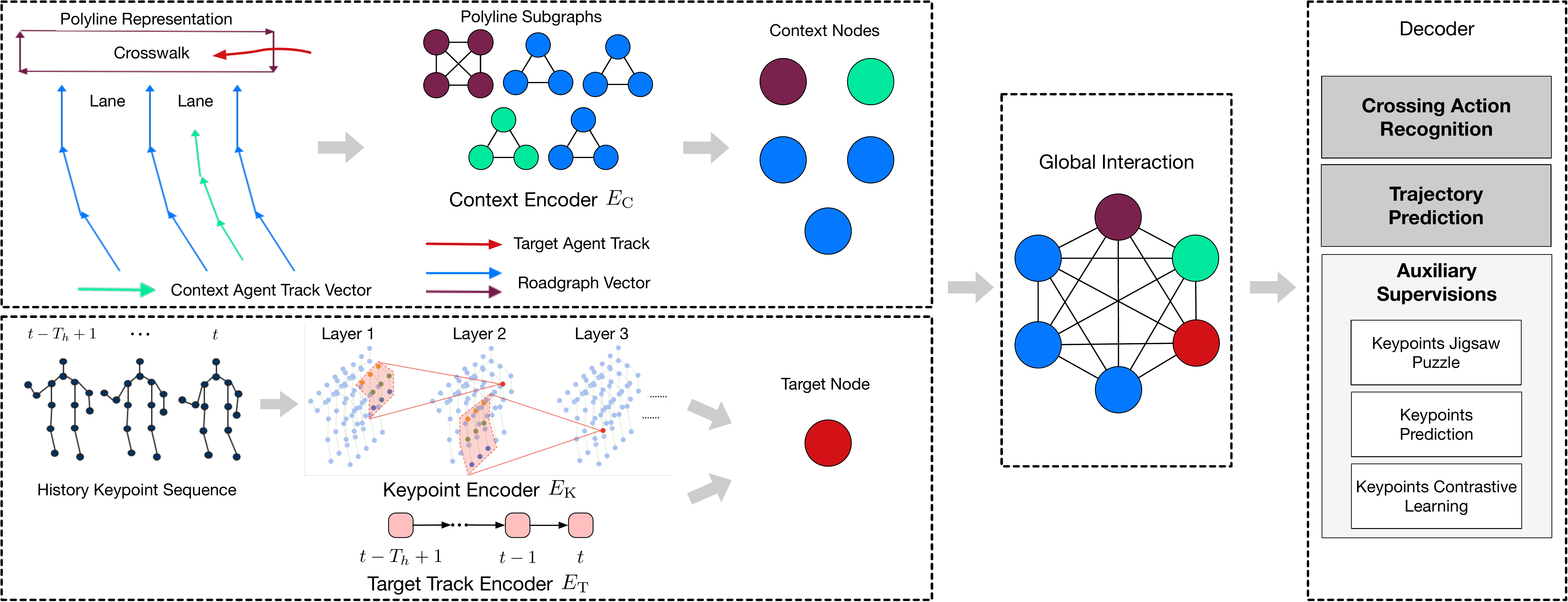}
	\caption{An illustrative diagram of the proposed method, which has an encoder-decoder architecture. More specifically, there are two encoding branches to extract features from context and keypoints information respectively and multiple decoding heads for major tasks and auxiliary supervisions.}
	\vspace{-0.5cm}
	\label{fig:framework}
\end{figure*}

\section{Problem Formulation}
We formulate the crossing action recognition as a binary classification task (i.e., crossing or non-crossing), and trajectory prediction as a regression task, which are integrated into a multi-objective optimization problem.
We denote $\mathbf{x}_{t-T_h+1:t}$ and $\mathbf{x}_{t+1:t+T_f}$ as the history and future trajectories of the target pedestrian respectively, where $t$ is the current frame, $\mathbf{x}_t = (x_t,y_t)$, and $T_h$ / $T_f$ are the history / prediction horizons. 
$\mathbf{k}_{t-T_h+1:t}$ / $\mathbf{k}_{t+1:t+T_f}$ denote the history / future human keypoints positions where $\mathbf{k}_t = \{(x^p_t,y^p_t,z^p_t), p=1,...,P\}$, $P$ is the number of keypoints in a human skeleton. $\mathbf{c}_t$ denotes the context agents and roadgraph information, which summarizes the information of a sequence of historical observations. 
The objective of our work is to infer the current crossing action $a_t$ and the future trajectory $\bar{\mathbf{x}}_{t+1:t+T_f}$ simultaneously based on all the history observations $\mathbf{x}_{t-T_h+1:t},\mathbf{k}_{t-T_h+1:t},\mathbf{c}_t$.

\section{Method}
\subsection{Model Overview}
A diagram of the multi-task learning framework is shown in Fig. \ref{fig:framework}, which has an encoder-decoder architecture.
The encoder consists of a context encoding channel $E_\text{C}$ which generates the embedding of roadgraph and context agents, a keypoints encoding channel $E_\text{K}$ which extracts the embedding of human pose and activity, and a target track encoding channel $E_\text{T}$ which extracts target pedestrian track features. 
The decoder consists of two individual heads for major tasks (i.e., crossing action recognition, trajectory prediction) and three auxiliary heads to enable auxiliary supervision for keypoints representation learning.

\subsection{Encoder}

\subsubsection{Context Encoder}
We adopt the vectorized representation proposed in \cite{gao2020vectornet} to represent the roadgraph and the historical trajectories of context traffic participants which may have interactions with the target pedestrian.
Specifically, the roadgraph (i.e., lanes, traffic signs) and trajectories are transformed into polylines with a variable number of vectors respectively. An illustration of polylines is shown in Fig. \ref{fig:framework}. Each polyline is used to construct a subgraph where each node represents a certain vector within the polyline. 

\subsubsection{Keypoints Encoder}
The goal of the keypoints encoder $E_\text{K}$ is to capture fine-grained information about human pose, motion, and activity.
In this work, a keypoints sequence is represented by 3D coordinates and visibility scores of all the human joints at each frame. 
Since the human skeleton can be naturally represented as a graph where nodes are human joints and edges are bones, we construct a spatio-temporal graph to represent a keypoints sequence where the same joint nodes at consecutive frames are linked by temporal edges. 
We adopt the spatial configuration partitioning strategy \cite{yan2018spatial} to obtain the adjacency matrix at each frame.
We employ the Spatio-Temporal Graph Convolutional Network with the same layer architecture in \cite{yan2018spatial} to extract both spatial and temporal patterns at different scales. 
A global average pooling layer is applied after graph convolutional layers to generate the final keypoints embedding.
More details can be found in \cite{yan2018spatial}.

\subsubsection{Target Track Encoder}
We use a recurrent neural network as the target track encoder $E_\text{T}$ to encode the history trajectory ($\mathbf{x}_{t-T_h+1:t}$) of the target pedestrian.
The extracted embedding is concatenated with the keypoints embedding to obtain the target node embedding.

\subsubsection{Global Interaction}
The polyline subgraphs and target node are used to construct a fully connected global interaction graph and message passing is applied to model the agent-agent and agent-road interactions between scene elements. We implement this global interaction graph as a self-attention layer, as in \cite{gao2020vectornet}. The output of the global interaction module is a global context embedding for each modeled agent while we only use that of the target pedestrian. 

\subsection{Decoder}
Our model has five decoding heads.
For each decoding head, besides using a complete embedding obtained by concatenating the context embedding and keypoints embedding as the input, we also apply the supervision directly on the keypoints embedding to enhance keypoints representation learning. The experimental results show improved performance brought by this co-training strategy.

\subsubsection{Crossing Action Recognition Head}
The crossing action recognition head is a fully-connected layer whose output are the probabilities of crossing/non-crossing actions.
We adopt a standard binary cross-entropy loss $\mathcal{L}_\text{AR}$ for training.

\subsubsection{Trajectory Prediction Head}
We adopt the Target-driveN Trajectory prediction method proposed in \cite{zhao2020tnt}, which consists of three stages:
(a) target prediction; (b) target-conditioned trajectory generation; and (c) trajectory scoring.

First, a set of $M$ target candidate points are sampled from a virtual grid centered on the predicted pedestrian and aligned with the pedestrian's heading based on the learned distribution over grids.
Second, the trajectory generator takes a target location and the embedding extracted by the encoder as input, and outputs a most likely future trajectory for the target pedestrian.
Finally, we estimate the likelihood of complete future trajectories and use a maximum entropy technique to score all the $M$ trajectories generated in the second stage.
A trajectory selection algorithm inspired by the non-maximum suppression is applied to reject near-duplicate trajectories to increase the diversity of final trajectory hypotheses.
Each stage has a corresponding loss term, which is combined to obtain the final loss function for the trajectory prediction head.
More details about loss design can be found in \cite{zhao2020tnt}.

\subsection{Auxiliary Supervisions}
In this section, we introduce three auxiliary supervisions on keypoints representation learning, which encourages the keypoints encoder to capture more fine-grained patterns of human pose dynamics and activities.
The keypoints encoder and the extracted keypoints embedding are shared across different auxiliary tasks.

\subsubsection{Keypoints Jigsaw Puzzle (KJP)}\label{sec:KJP}
Different from the image jigsaw puzzle where image patches are shuffled into random spatial configurations, the goal of solving keypoints jigsaw puzzle is to identify the correct permutation of a given keypoints sequence in which subsequences are randomly shuffled, which encourages the keypoints encoder to learn temporal relations among different segments.

\begin{figure}[!tbp]
	\centering
	\includegraphics[width=0.89\columnwidth]{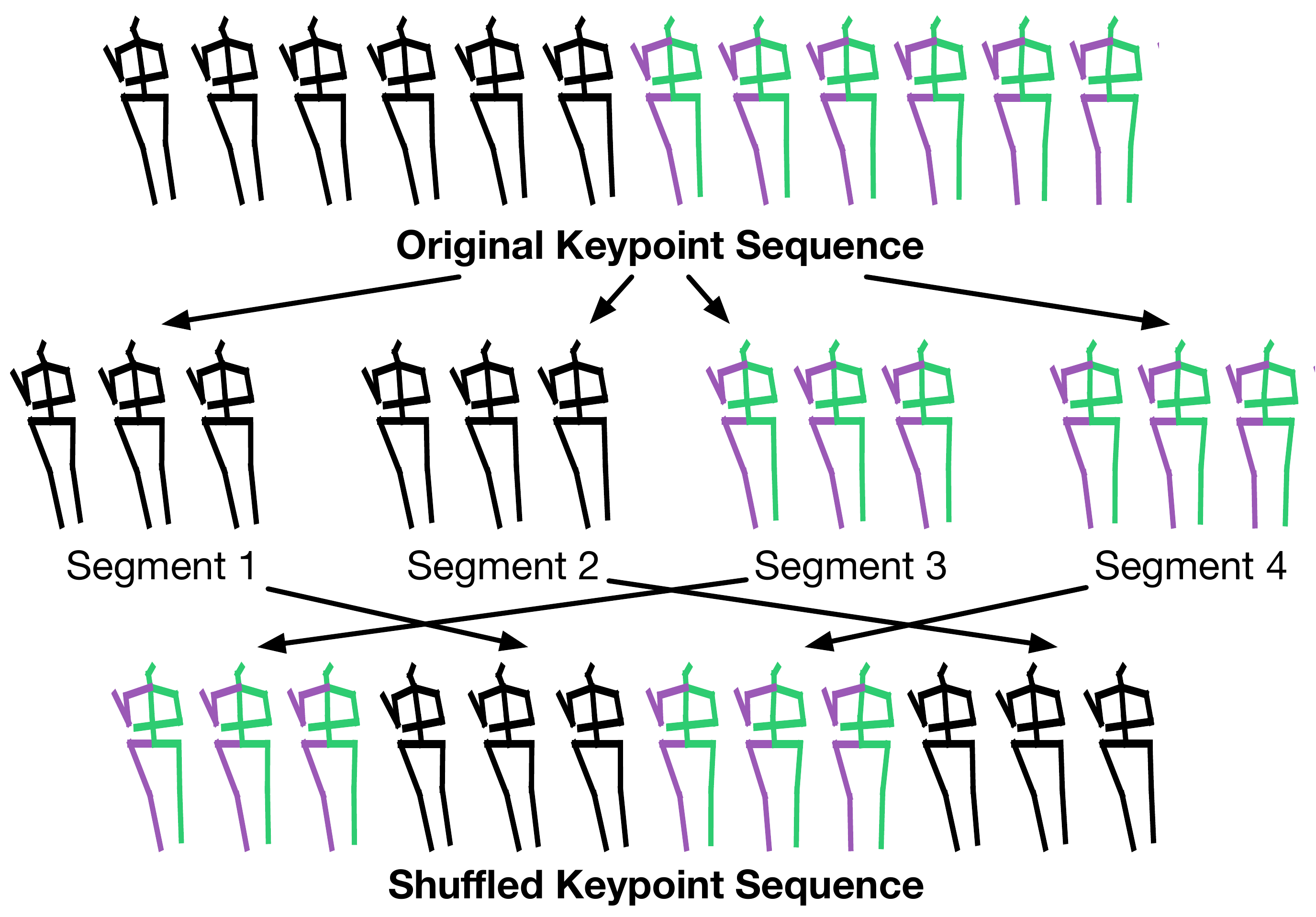}
	\caption{A diagram of the shuffled sequence generation process for keypoints jigsaw puzzle.}
	\vspace{-0.5cm}
	\label{fig:KJP}
\end{figure}

A diagram of the jigsaw puzzle sequence generation process is shown in Fig. \ref{fig:KJP}. 
The original keypoints sequence $\mathbf{k}_{t-T_h+1:t}$ is divided into $S$ segments with an equal length of $T_h/S$ frames.
Then we randomly shuffle these segments to obtain a new sequence which can lead to $S!$ permutations. 
Instead of shuffling the original coordinates of keypoints sequences, we fix the center location of human skeletons at each frame and only shuffle the relative coordinates of keypoints with respect to the skeleton center over different frames. 
This can encourage the model to capture subtle patterns of human pose dynamics and avoid the trivial shortcut to only capturing the change of center locations for the inference of the correct permutation.
We formulate this auxiliary task as a classification problem with $S!$ classes, and the KJP head $h_\text{KJP}(\cdot)$ is trained to infer the probability of each class (i.e., permutation) for a shuffled input sequence.
We use a standard cross-entropy loss computed by
\begin{equation}
	\mathcal{L}_\text{KJP} = - \frac{1}{N}\sum_{i=1}^{N} y_i \log h_\text{KJP}(E_\text{k}(\tilde{\mathbf{k}}^{i}_{t-T_h+1:t})),
\end{equation}
where $\tilde{\mathbf{k}}^{i}_{t-T_h+1:t}$ and $y_i$ denote a shuffled keypoints sequence and its corresponding one-hot permutation label, respectively.
$N$ denotes the total number of pedestrians.

\subsubsection{Keypoints Prediction (KP)}
Keypoints prediction aims to infer the future keypoints locations based on a sequence of historical observations, which requires a more fine-grained understanding of human pose dynamics than trajectory prediction. 
The keypoints prediction head $h_\text{KP}(\cdot)$ is a multi-layer perceptron, which takes in the keypoints embedding as input and outputs the future keypoints locations.
We adopt the mean square error loss computed by
\begin{equation}
	\mathcal{L}_\text{KP} = \frac{1}{NT_f}\sum_{i=1}^{N} \sum_{t'=t+1}^{t+T_f} ||\mathbf{k}^{i}_{t'}-\hat{\mathbf{k}}^{i}_{t'} ||^2,
\end{equation}
where $\hat{\mathbf{k}}^{i}_{t'}$ is the predicted keypoints coordinates at time $t'$.

\subsubsection{Keypoints Contrastive Learning (KCL)}
\begin{figure}[!tbp]
	\centering
	\includegraphics[width=0.94\columnwidth]{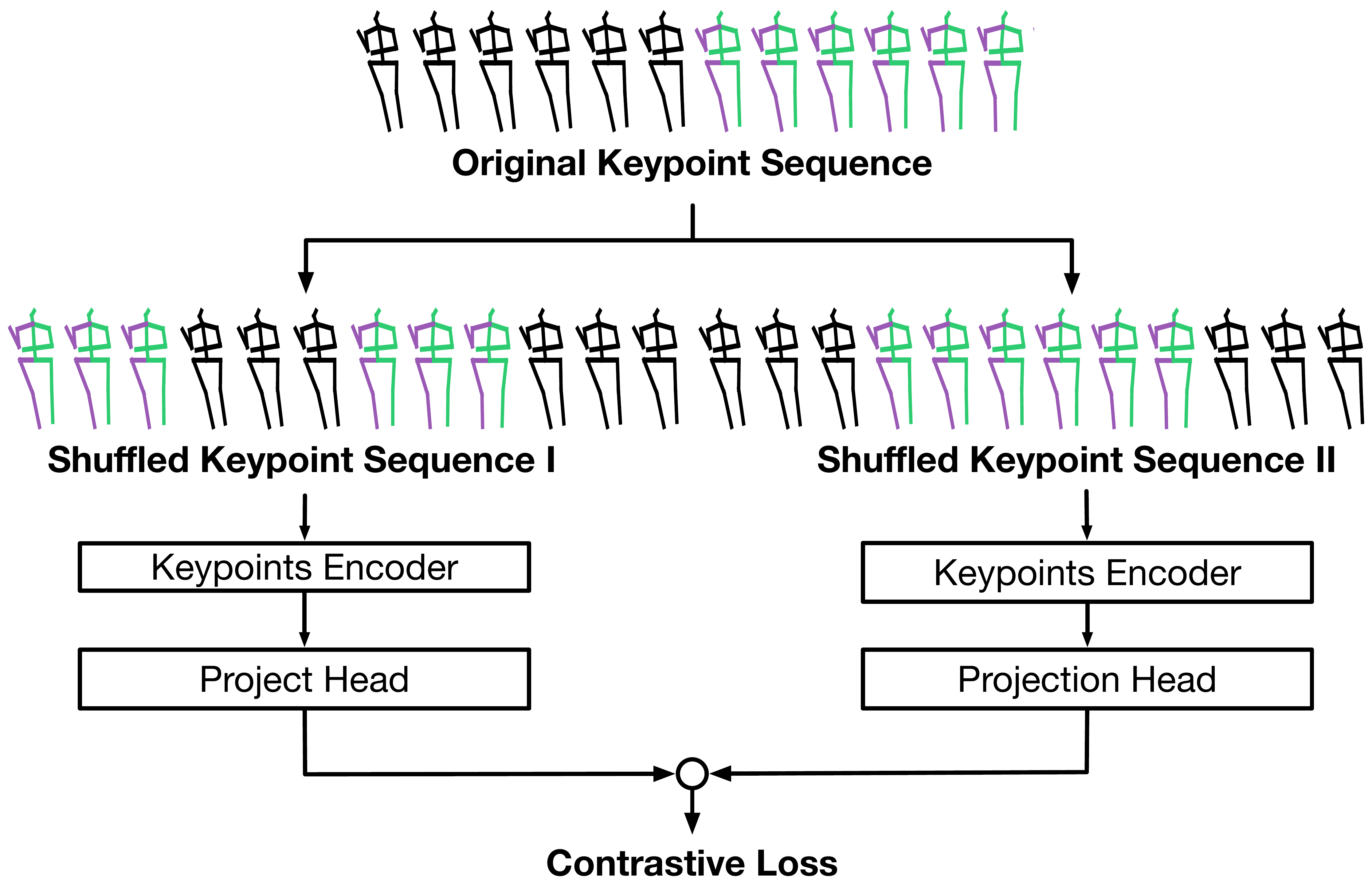}
	\caption{A diagram of the auxiliary contrastive learning task leveraging the shuffled sequences in the jigsaw puzzle.}
	\vspace{-0.5cm}
	\label{fig:KCL}
\end{figure}

As an analogy to learning similarity between transformed images by contrastive learning, we propose to adapt the technique to keypoints sequences as an auxiliary supervision to improve representation learning by learning high-level temporal similarity among keypoints sequences.
A diagram of the auxiliary contrastive learning is shown in Fig. \ref{fig:KCL}.

We employ a similar approach to SimCLR \cite{chen2020simple} to discover the cosine similarity between the keypoints sequences transformed from the same original sequence. 
In this work, we use the same shuffling strategy as described in Section \ref{sec:KJP} to obtain transformed sequences.  
The keypoints encoder takes in each transformed sequence and outputs a corresponding embedding. A projection head $h_\text{KCL}$ maps the keypoints embedding to a feature space.

More specifically, we randomly sample a mini-batch of $N_b$ keypoints sequences and define the contrastive prediction task on pairs of the augmented sequences derived from the mini-batch, which results in $2N_b$ keypoints sequences.
Given a positive pair, the other $2(N_b-1)$ transformed sequences are treated as negative examples.
The cosine similarity between vectors $\mathbf{u}$ and $\mathbf{v}$ is defined as $\cos(\mathbf{u},\mathbf{v}) = \mathbf{u}^\top\mathbf{v}/||\mathbf{u}|||\mathbf{v}||$.
Then the keypoints contrastive loss between pair $(p,q)$ is
\begin{equation}
	\mathcal{L}_\text{KCL} = -\log \frac{\exp(\beta \cos(\mathbf{z}_p,\mathbf{z}_q))}{\sum_{r\neq p}\exp(\beta \cos(\mathbf{z}_p,\mathbf{z}_r))},
\end{equation}
where $\mathbf{z}_p = h_\text{KCL}(E_\text{k}(\tilde{\mathbf{k}}^p_{t-T_h+1:t}))$ and $\beta$ is temperature.

\subsection{Loss Functions and Training}
Since the proposed framework deals with multiple tasks simultaneously, the complete loss function is a weighted sum of all the loss components brought by each head.
More specifically, the final loss is computed by
\begin{equation}
	\mathcal{L} = \lambda_\text{AR}\mathcal{L}_\text{AR} + \lambda_\text{TP}\mathcal{L}_\text{TP} + \lambda_\text{KJP}\mathcal{L}_\text{KJP} + \lambda_\text{KP}\mathcal{L}_\text{KP} + \lambda_\text{KCL}\mathcal{L}_\text{KCL},
\end{equation}
where $\lambda_\text{AR}$, $\lambda_\text{TP}$, $\lambda_\text{KJP}$, $\lambda_\text{KP}$ and $\lambda_\text{KCL}$ are loss weights which are determined empirically.

\section{Experiments}

\setlength{\tabcolsep}{2pt} 
\begin{table*}[!tbp]
	\fontsize{8}{9}\selectfont
	\begin{center}
		\resizebox{\textwidth}{!}{
			\begin{tabular}{m{4.3cm}<{\centering}|m{1.8cm}<{\centering} m{1.8cm}<{\centering} m{1.8cm}<{\centering} m{1.8cm}<{\centering}| m{2cm}<{\centering} m{2cm}<{\centering}}
				\toprule
				\midrule
				\multirow{4}*{\shortstack[lb]{}} 
				& \multicolumn{4}{c|}{Crossing Action Recognition (AR)} & \multicolumn{2}{c}{Trajectory Prediction (TP)} \\
				\cline{2-7}
				& & & & & & \\[-0.1cm]
				Method & Acc $\uparrow$ & AUC$_\text{PR}$ $\uparrow$ & F1 $\uparrow$ & Prec $\uparrow$ & 4.0s minADE$_6$ $\downarrow$ & 4.0s minFDE$_6$ $\downarrow$ \\
				\midrule 
				TNT \cite{zhao2020tnt} & -- & -- & -- & -- & 0.404 & 0.703\\
				TNT+AR  & 0.959 & 0.957 & 0.933 & 0.893 & 0.418 & 0.733\\
				\midrule
				Ours (AR)  & 0.960 & 0.957 & 0.935 &  0.900 & -- & -- \\
				Ours (TP)  & -- & -- & -- & -- & 0.397 & 0.704 \\
				Ours (AR+TP)  & 0.963 & 0.965 & 0.940 & 0.900 & 0.382 & 0.682 \\
				\midrule
				Ours (AR+TP+KJP)  & 0.979 & 0.983 & 0.964 & 0.937 & 0.392 & 0.685\\
				Ours (AR+TP+KJP+KP)  & 0.978 & 0.983 & 0.964 & 0.936 & 0.384 & 0.665\\
				Ours (AR+TP+KJP+KP+KCL)  & \textbf{0.980} & \textbf{0.984} & \textbf{0.966} & \textbf{0.942} & \textbf{0.373} & \textbf{0.638}\\
				\bottomrule
			\end{tabular}
		}
	\end{center}
	\vspace{-0.2cm}
	\caption{Model performance on the in-house dataset. $\uparrow$/$\downarrow$ indicate greater/smaller values are better. AR: action recognition; TP: trajectory prediction; KJP: keypoints jigsaw puzzles; KP: keypoints prediction; KCL: keypoints contrastive learning.}
	\vspace{-0.4cm}
	\label{tab:IH_table}
\end{table*}

\subsection{Datasets}
We validated the proposed method on two real-world driving datasets: an in-house dataset and Waymo Open Dataset (Perception) \cite{sun2020scalability}. 
Both datasets provide 3D object bounding boxes and LiDAR point clouds around the autonomous vehicle. The 3D bounding boxes in the in-house dataset are generated by an in-house perception pipeline, and the 3D bounding boxes in Waymo Open Dataset (WOD) are human-labeled. We use 2.0 seconds of history sampled at 10Hz (20 frames), and predict trajectory in the future 4.0 seconds sampled at 2Hz (8 frames).
The in-house dataset also provides detailed roadgraph information and human labels on the crossing action of pedestrians.
The in-house dataset provides 2,039,520 training examples (1,451,460 negatives, 588,060 positives) and 918,049 testing samples (642,523 negatives and 275,526 positives).
Since WOD does not have crossing action labels, we removed the crossing action recognition branch of our proposed models and only trained and evaluated the trajectory prediction task. Since there is no roadgraph information in WOD, we only encoded context agent tracks in the context embedding.
The WOD dataset provides 229,745 training samples and 51,945 testing samples.
We used the pose estimation model in \cite{keypoint_cvpr_submission} to infer 3D human keypoints from laser points for both datasets. 

\subsection{Evaluation Metrics}
For crossing action recognition, we adopt widely used binary classification metrics as in \cite{yau2021graph}: accuracy (Acc), area under the precision-recall curve (AUC$_\text{PR}$), F1 score (F1) and precision (Prec).
For trajectory prediction, we adopt the standard distance-based metrics as in \cite{zhao2020tnt}: minimum average displacement error (minADE$_k$) and minimum final displacement error (minFDE$_k$) where $k$ is the number of trajectory hypotheses. 

\subsection{Implementation Details}
The same architectures of model backbones and hyperparameters in \cite{gao2020vectornet}, \cite{yan2018spatial}, and \cite{zhao2020tnt} are adopted as our context encoder, keypoints encoder, and trajectory decoder, respectively.
The crossing action head is a fully connected layer.
The keypoints jigsaw puzzle head is a fully connected layer.
The projection head in keypoints contrastive learning is a three-layer MLP with hidden size = 64. For loss weights, we used $\lambda_\text{CA} = 1.0$, $\lambda_\text{TP} = 1.0$, $\lambda_\text{KJP} = 0.01$, $\lambda_\text{KP} = 0.05$ and $\lambda_\text{KCL} = 0.0001$.
For KJP, we used $S=4$ segments and a fully-connected layer that outputs the probability of $S!$ classes. For the KCL head, we used temperature parameter $\beta=1.0$ and the projection head is a three-layer MLP with hidden size = 64.
A batch size of 512 was used and the models were trained for 15 epochs with early stopping using the Adam optimizer with an initial learning rate of 0.01. We train for 15 epochs including learning rate warm up (linearly increasing to initial learning rate) in the first 6 epochs, and decay learning rate by 0.9 every 6 epochs. 
Our method achieves an inference time of 4.7ms for a batch of 32 pedestrians on an NVIDIA V100 GPU, which is sufficient for running onboard.

\subsection{Quantitative and Ablative Analysis}
In this section, we provide a comprehensive quantitative and ablative analysis to demonstrate the effectiveness of each component in our framework. 
The quantitative experimental results are shown in Table \ref{tab:IH_table} and Table \ref{tab:WOD_internal_noroad_table}. 
Our complete model (AR+TP+KJP+KP+KCL for the in-house dataset, TP+KJP+KP+KCL for WOD) achieves the best performance consistently on both datasets.

\vspace{0.1cm}
\noindent
\textbf{Keypoints information}
To allow for a fair comparison, we add a crossing action recognition head to the TNT baseline.
Comparing TNT+AR and Ours (AR+TP), we see that utilizing 3D human keypoints can improve the performance of both major tasks. 
With the human keypoints information, the model can capture richer cues about pedestrians' subtle motions and activities in addition to trajectory patterns.

\setlength{\tabcolsep}{2pt} 
\begin{table}[!tbp]
	\fontsize{8}{8.}\selectfont
	\begin{center}
		\resizebox{0.9\columnwidth}{!}
		{\begin{tabular}{c|c c}
				\toprule
				\midrule
				\multirow{2}{*}{Method} & \multicolumn{2}{c}{Trajectory Prediction (TP)}\\ 
				\cline{2-3}
				& &  \\[-0.0cm]
				& 4.0s minADE$_6$ $\downarrow$ & 4.0s minFDE$_6$ $\downarrow$ \\
				\midrule
				TNT \cite{zhao2020tnt} & 0.249 & 0.494\\
				\midrule
				Ours (TP) & 0.239 & 0.451   \\
				\midrule
				Ours (TP+KJP+KP+KCL) & \textbf{0.237}&\textbf{0.442} \\
				\bottomrule
		\end{tabular}}
	\end{center}
	\vspace{-0.2cm}
	\caption{Model performance on the Waymo Open Dataset.}
	\vspace{-0.6cm}
	\label{tab:WOD_internal_noroad_table}
\end{table}

\vspace{0.1cm}
\noindent
\textbf{Co-learning of crossing action recognition and trajectory prediction}
The two major tasks are highly correlated, which require the model to extract both high-level motion patterns and fine-grained cues about human pose and activity.
The results show that training the model on both tasks jointly achieves better performance than learning from an individual task.
A potential reason is that action recognition focuses more on learning the road context and keypoints features while trajectory prediction focuses more on learning history motions and the interactions between target pedestrian and context agents, which complement and enhance each other.

\vspace{0.1cm}
\noindent
\textbf{Auxiliary supervisions}
We conducted an ablation study on auxiliary supervisions for keypoints representation learning.
Comparing Ours (AR+TP) and Ours (AR+TP+KJP), we see that the performance of crossing action recognition improves by a large margin consistently, which implies the effectiveness of learning temporal relations in keypoints sequences.
By adding the supervision from the keypoints prediction task, we can see further improvement in Ours (AR+TP+KJP+KP) for both tasks, especially in trajectory prediction since KP encourages the model to learn both macro motions (i.e., location) and micro dynamics (i.e., pose and activity) of the pedestrians.
Finally, in Ours (AR+TP+KJP+KP+KCL) the keypoints contrastive learning leads to higher accuracy by learning inherent representations and similarity between keypoints sequences.

\begin{figure}[!tbp]
	\centering
	\includegraphics[width=\columnwidth]{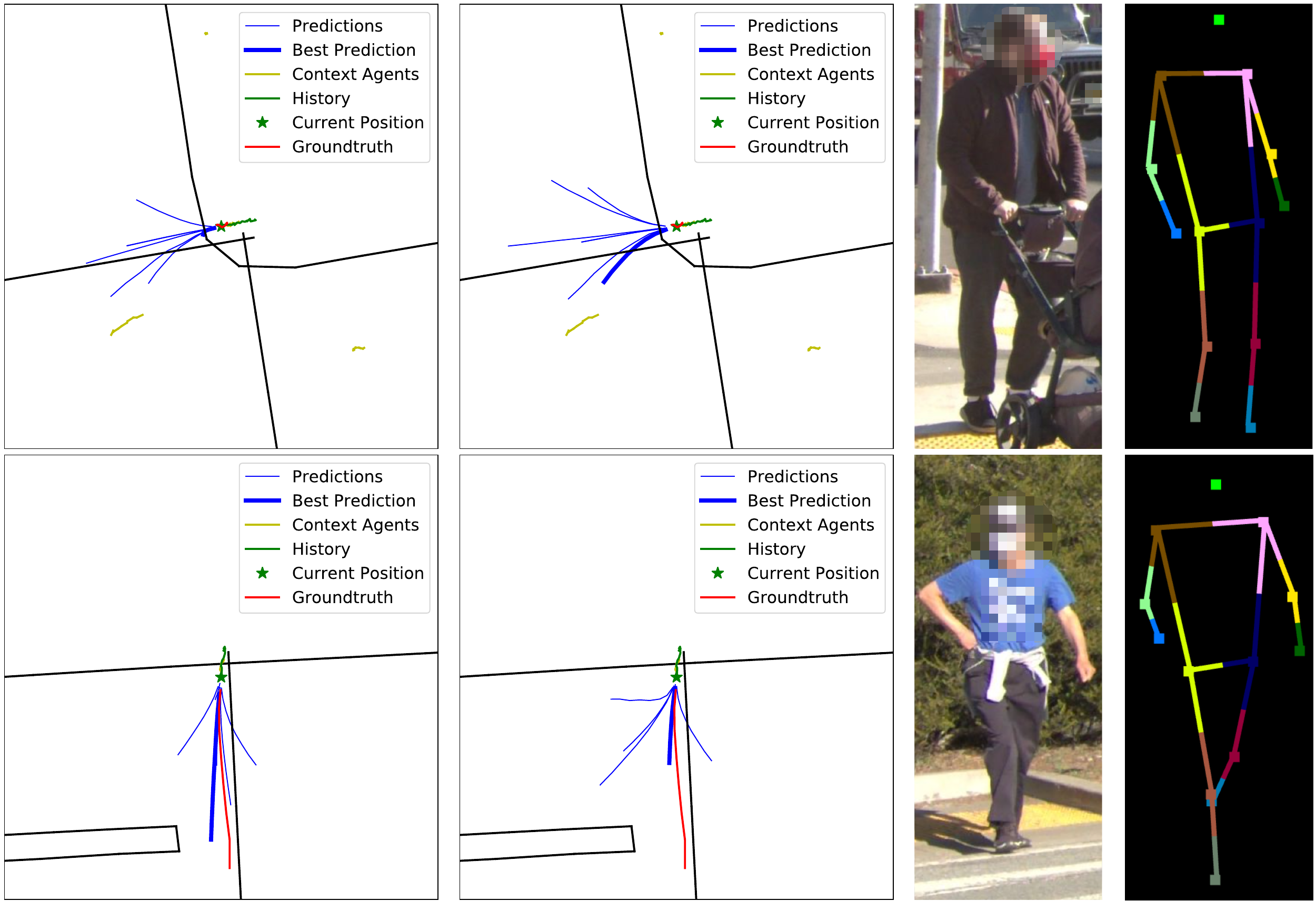}
	\caption{Visualization of the predicted trajectories. Each row corresponds to one example, and four columns correspond to: 1) trajectory prediction of our model with keypoints; 2) trajectory of baseline TNT model; 3) camera patch; 4) inferred keypoints of the pedestrian. In the first example, the pedestrian approaches an intersection and waits for the traffic, and the keypoints provide the information that the pedestrian is not in a walking pose with two legs straight. Our model successfully predicts an almost stationary trajectory (the short bold blue line), while the baseline consistently predicts the pedestrian to proceed forward. In the example at the second row, the pedestrian speeds up and starts to walk on the crosswalk, and the keypoints show a walking pose thus our model predicts longer trajectories moving forward.}
	\vspace{-0.3cm}
	\label{fig:trajectories_visualization}
\end{figure}

\vspace{0.1cm}
\noindent
\textbf{Trajectory prediction on WOD}
The trajectory prediction metrics of all methods are better on WOD than on the in-house dataset, because human-labeled bounding boxes are used for history/future trajectories in WOD experiments, which means both the model inputs and trajectory supervision have smaller error and noise compared to using the bounding boxes inferred by the model. Compared with the strong baseline TNT, keypoints information and auxiliary supervisions improve trajectory prediction performance.   

\subsection{Qualitative Analysis}

\begin{figure}[!tbp]
	\centering
	{
		{\includegraphics[width=\columnwidth]{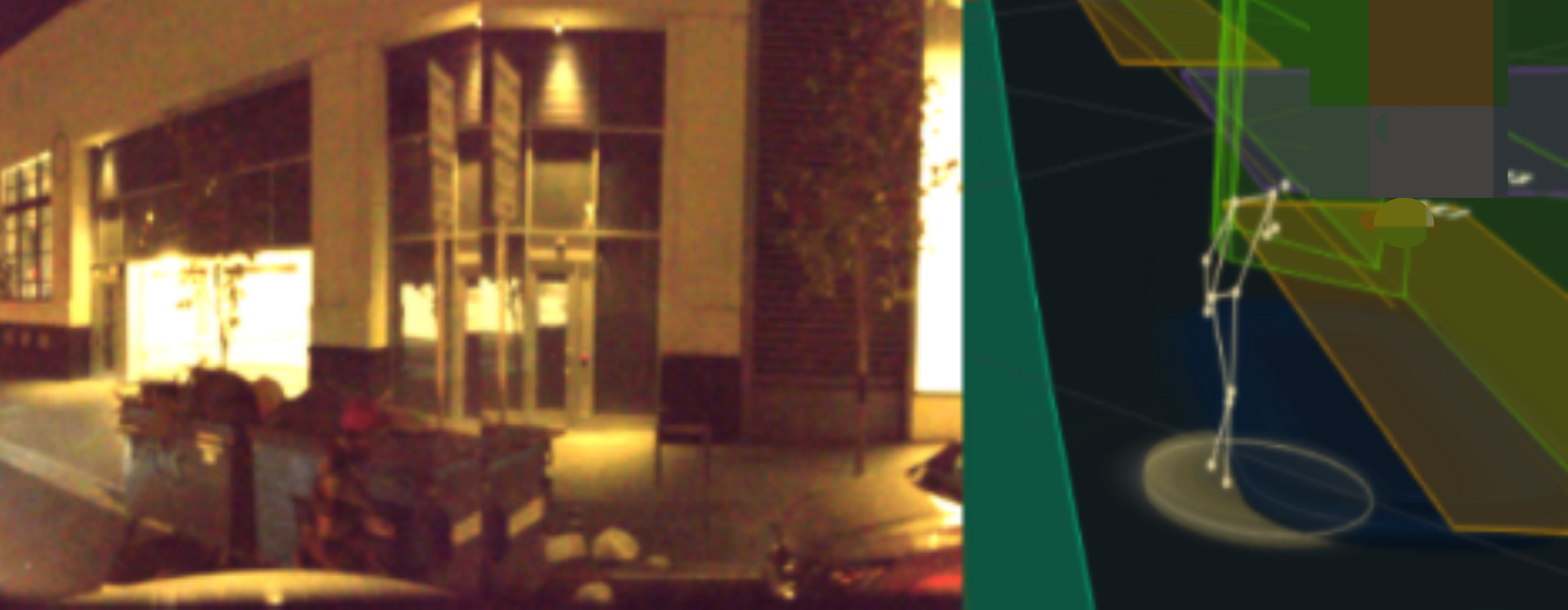}}
		\subcaption[width=\columnwidth]{Crossing prob.: 0.521 (w/o Keypoints) $\rightarrow$ 0.203 (w/ Keypoints) }
		{\includegraphics[width=\columnwidth]{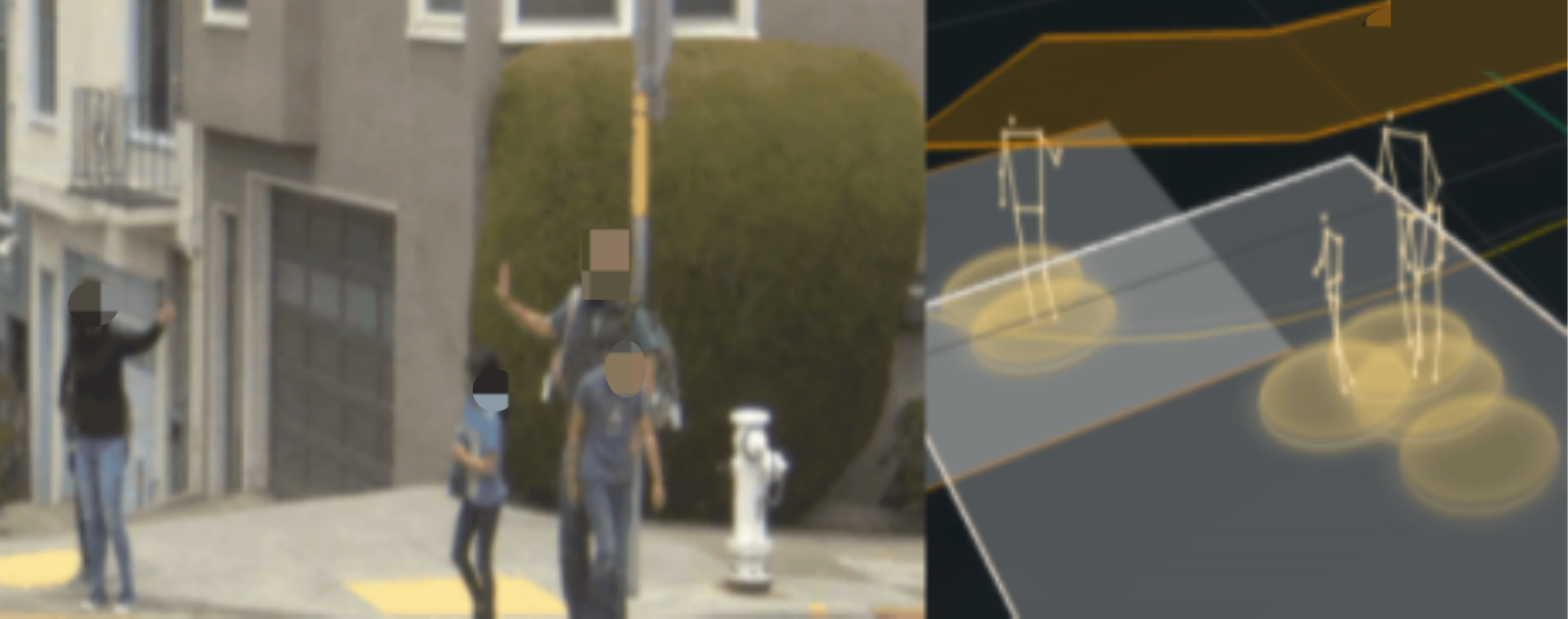}}
		\subcaption[width=\columnwidth]{Crossing prob.: 0.288 (w/o Keypoints) $\rightarrow$ 0.608 (w/ Keypoints) }
	}
	\caption{The visualization of testing cases and a comparison between the models with/without keypoints information.}
	\vspace{-0.1cm}
	\label{fig:examples}
\end{figure}

We visualize the predicted trajectories in Fig. \ref{fig:trajectories_visualization}. It shows that our model can generate accurate and diverse trajectory hypotheses. 
Our model, equipped with a strong keypoints encoder, is able to capture the action and pose cues of the pedestrians.
We also provide two typical examples in Fig. \ref{fig:examples} where utilizing human keypoints information enables the model to capture more subtle patterns of pedestrians' pose and activity for crossing action recognition.
More specifically, in (a) the model w/ keypoints predicts no crossing action by utilizing the pose information when the pedestrian is bending down to pick up stuff even though he/she is on the road. This activity cannot be captured by the historical trajectory and context embeddings extracted by the context encoder. Without the keypoints information, the model tends to heavily depend on the on-road / off-road state of pedestrians to determine their crossing actions.
Reducing false alarms on crossing action can help develop assertive but safe driving strategies for autonomous vehicles. 
In (b), the model w/ keypoints accurately predicts positive crossing action as soon as the pedestrian stops waving and starts turning to the crossing direction. These activities can be detected from the motion of the human skeleton earlier than from the trajectory. The early detection of crossing action helps autonomous vehicles reduce collision risks.

\section{CONCLUSIONS}
In this paper, we demonstrated the first effective utilization of 3D human keypoints information for pedestrian crossing action recognition and trajectory prediction in a unified multi-task learning framework. Our approach captures fine-grained information of human pose, motion, and activity, while being efficient to run onboard an autonomous vehicle.
To improve keypoints representation learning, we also proposed to apply auxiliary supervisions which further enhance the model performance on the two primary tasks.
The framework is validated on an in-house dataset and a public benchmark dataset. 
The results show that the co-learning of two majors tasks outperforms learning individually, and the model with keypoints information performs better especially in capturing special human activities and detecting crossing action early for the pedestrians near the road edge.

\printbibliography

\addtolength{\textheight}{-12cm}   

\end{document}